\documentclass[11pt]{article}

\usepackage[preprint]{acl}

\usepackage{times}
\usepackage{latexsym}
\usepackage{amsmath}
\usepackage{amssymb}
\usepackage{bbm}

\usepackage[T1]{fontenc}

\usepackage[utf8]{inputenc}
\usepackage{booktabs}

\usepackage{microtype}

\usepackage{inconsolata}
\usepackage{xcolor}

\usepackage{graphicx}

\title{Represented but Not Used: Value–Policy Dissociation in LLM Agents}

\author{
  Yuzhang Luo \\
  Peking University \\
  \texttt{luoyuzzhang@stu.pku.edu.cn}
  \And
  Mohamed Aghzal \\
  George Mason University \\
  \texttt{maghzal@gmu.edu}
  \AND
  Liangming Pan \\
  Peking University \\
  \texttt{liangmingpan@pku.edu.cn}
  \And
  Ziyu Yao\thanks{Corresponding author.} \\
  George Mason University \\
  \texttt{ziyuyao@gmu.edu}
}

\begin{document}
\newcommand{\mo}[1]{\textcolor{blue}{[Mo: #1]}}
\newcommand{\moedit}[1]{\textcolor{blue}{#1}}

\newcommand{\yuzhang}[1]{\textcolor{red}{[Yuzhang: #1]}}
\newcommand{\yuzhangedit}[1]{\textcolor{red}{#1}}

\newcommand{\zyc}[1]{\textcolor{orange}{[ZY: #1]}}
\newcommand{\zyedit}[1]{\textcolor{orange}{#1}}

\maketitle

\begin{abstract}
Large language model (LLM) agents need to choose actions whose consequences unfold over multiple future steps, yet poor decisions do not reveal whether the model failed to represent those consequences or failed to use them. Mechanistic interpretability has studied internal representations of environment state and plans, but sequential decision making also requires evaluating how candidate actions change future outcomes. Reinforcement learning formalizes this as \emph{action value}. We study this computation in Qwen2.5-7B-Instruct, Qwen2.5-32B-Instruct, and GPT-OSS-20B across a partially observed Grid environment and WebShop, a simulated web shopping environment. We find that the relative values of candidate actions are linearly decodable from intermediate hidden states, including at states where the policy prefers a worse action. We call this gap between value representation and policy choice \emph{value–policy dissociation}. To investigate how represented value relates to action selection, we compare direct interventions along the value-readout direction with \emph{Natural Value Transport (NVT)}, a transformation learned from hidden-state changes that naturally result from differences in action value. While readout-direction interventions have limited and inconsistent effects, NVT more consistently influences action preferences, particularly at later layers. We also test a closed-loop controller that uses the value readout as a sensor and NVT as an actuator, observing improvements in task outcomes. These results suggest that LLM agents can represent useful prospective value without reliably translating it into policy choice.\footnote{The source code and data will be released.}

\end{abstract}

\section{Introduction}

\begin{figure*}[!t]
    \centering
    \includegraphics[width=\textwidth]{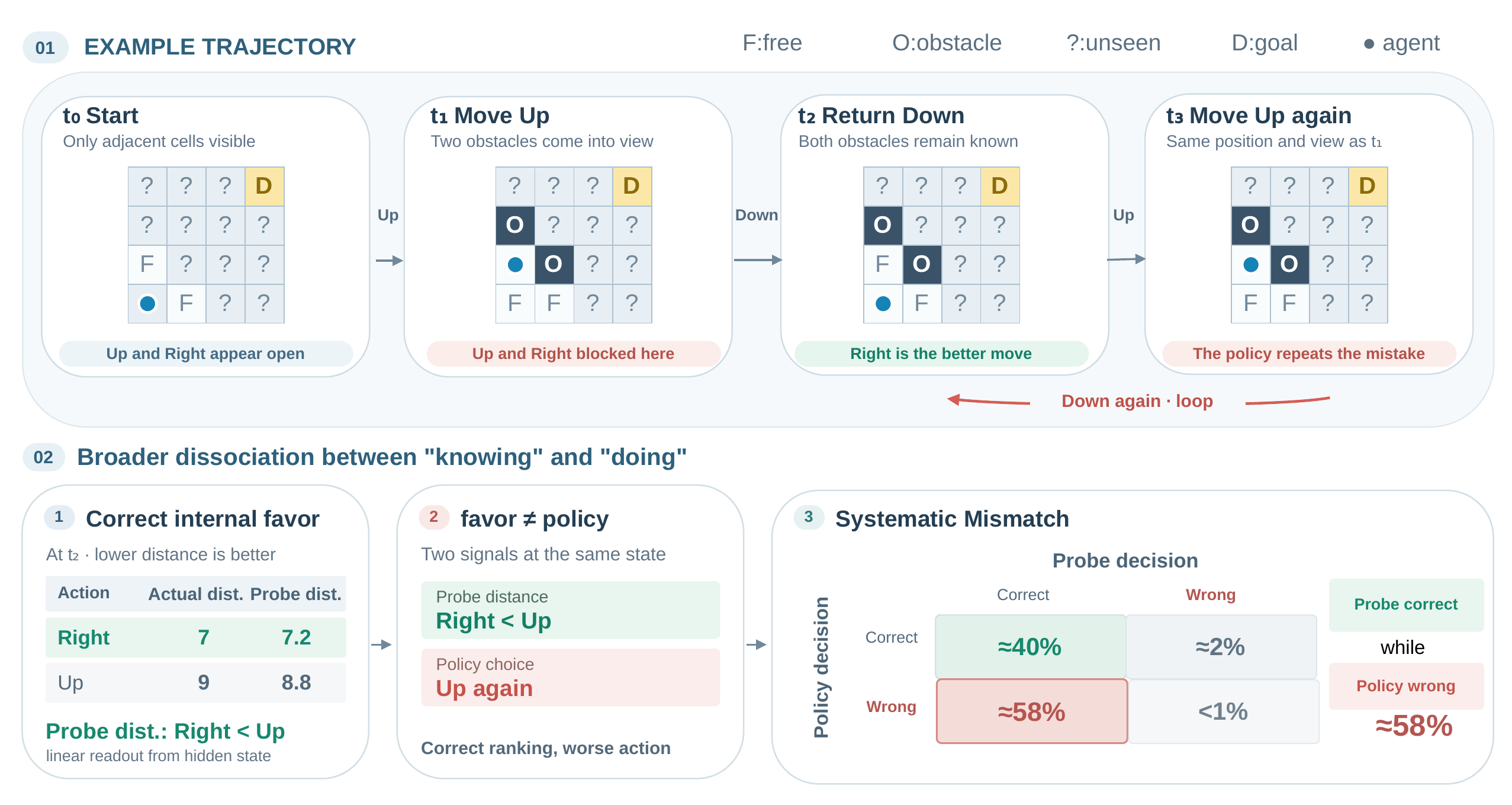}
    \caption{
Value–policy dissociation from a motivating example to a systematic pattern. Top: a looping Grid trajectory in which the agent repeats an inferior action despite observations favoring a better alternative. Bottom: Generalizing to broader scenarios, a linear readout of the hidden state favors the better action while the policy selects the worse one; this probe–policy mismatch recurs across states.
    }
    \label{fig:overview}
\end{figure*}

Recent years have seen rapid progress in the ability of large language models (LLMs) to act in interactive environments, where models must choose actions over multiple steps to obtain delayed outcomes \citep{hao-etal-2023-reasoning,aghzal2024look,xia-etal-2025-agentrm,DBLP:conf/aaai/ZhaiYXFYDW25}. Despite these advances, LLM agents still struggle with planning tasks that require sustained progress toward a goal \citep{ICLR2024_e9df36b2,aghzal2024look,wang2025hell,aghzal-etal-2026-llm}.
A typical example is \emph{looping}: agents revisit previously visited states or repeat earlier actions without making progress, even when the environment has revealed information that should change their next move \citep{ICLR2024_4410c071,chung2025evaluating}. While looping is one salient behavior, it also points to a broader failure mode in sequential decision making: agents may choose an inferior action even when the information they have already observed clearly favors a better alternative. This raises a basic question about the internal computations that support sequential decision making in these models.


Mechanistic interpretability (MI) has begun to uncover structured internal representations of environment state and world structure in sequence models and LLM agents. In Othello, \citet{li2023emergent} identified representations of board state, while \citet{nanda-etal-2023-emergent} showed that board-state information is linearly represented and can be causally manipulated to affect move predictions. Recent studies further show that these models encode spatial relations \citep{ICLR2024_0a605985,karvonen2024emergent,martorell2025text}, temporal dynamics and task-relevant trajectories \citep{NEURIPS2024_37d9f191}, and a cognitive map of the environment state \citep{arghal2026a}. 
However, these studies have largely focused on \emph{what models represent about the world state}, which does not directly explain the models' decision making. Indeed, in our preliminary experiments, we found that agents internally encoded rather precise state representations when they exhibited the looping behavior.

In this work, we extend this line of analysis beyond world-state representation to examine \emph{how models evaluate a prospective action} and, crucially, its connection to policy choice. In reinforcement learning, this evaluation is formalized as \emph{action value}, which summarizes the expected future return of a candidate action \citep{sutton1991dyna,ha2018world,schrittwieser2020mastering}. For LLM agents, however, it remains unclear whether such prospective evaluations are internally represented during sequential decision making.
{We therefore ask two questions: (Q1) When an agent selects an inferior action, does it fail to accurately represent the values of available alternatives? (Q2) If these values are represented correctly, how do they relate to action selection, and can we leverage them to improve the agent's decisions?}

We study these questions in Qwen2.5-7B-Instruct, Qwen2.5-32B-Instruct \citep{DBLP:journals/corr/abs-2412-15115}, and GPT-OSS-20B \citep{agarwal2025gpt} across a partially observed Grid environment and WebShop \citep{NEURIPS2022_82ad13ec}. Figure~\ref{fig:overview} illustrates our analysis pipeline. We probe hidden activations to examine whether prospective action value is represented when the policy selects a worse action. Across both environments, action value information is linearly decodable from intermediate hidden states even when the policy makes suboptimal choices, revealing a dissociation between value representation and policy selection.
We then investigate how represented value information can influence action choice. Direct steering along the value-readout direction has limited effects, whereas hidden-state changes naturally associated with value variations identify a later policy-relevant transformation, {which we call Natural Value Transport (NVT). Using the value readout as a diagnostic signal and NVT as a control mechanism, we improve agent decisions in both environments. For example, value-guided steering increases WebShop reward of Qwen2.5-32B-Instruct from 0.358 to 0.452.} These results show that LLM agents can represent prospective value without reliably using it for decisions, and that identifying value-associated activation directions that causally influence policy preferences provides a practical mechanism for improving agent behavior.


\section{Preliminaries} 
\label{sec:preliminaries}

\subsection{Formulation of LLM Agents} 
\label{sec:agent-formulation}

An LLM agent interacts with an environment over multiple steps. We formulate this interaction as a partially observable Markov decision process (POMDP), where the agent acts based on its observation history rather than directly observing the underlying environment state. At step \(t\), the agent receives the current observation together with the interaction history, which defines a decision state \(s_t\). The model forms a hidden representation \(h_\ell(s_t)\) at layer \(\ell\) and induces a policy
\(\pi(a\mid s_t)\) over the available actions \(a\in\mathcal{A}(s_t)\). After an action is selected, the environment returns a new observation, which is appended to the interaction history for the next decision. Sequential decision making therefore requires the agent to use information accumulated across the interaction history to make progress toward a delayed goal. In particular, the agent must maintain relevant information about the environment, anticipate how different actions affect future outcomes, and use this information to guide its next action choice.

We study this interaction process in two environments that place different demands on sequential decision making: a partially observed navigation task with a small, fixed action space, and a web interaction task with a larger, state-dependent action space, which we introduce next. 

\subsection{Experimental Setup}
\label{sec:tasks}


\paragraph{Models.} We study three models in total: two scales of the same instruction-tuned model family, Qwen2.5-7B-Instruct and Qwen2.5-32B-Instruct~\citep{DBLP:journals/corr/abs-2412-15115}, and GPT-OSS-20B~\citep{agarwal2025gpt}. 
The models are prompted to follow an observation-action loop: they receive the current observation and relevant interaction history and then generate a single action at each step. For concrete prompts and interaction examples, please refer to Appendix~\ref{app:prompts}.

\paragraph{Environments.}
We evaluate two environments. In \textbf{Grid}, we collect 1,400 maps on a \(5\times 5\) grid.
At each step, the agent chooses one of four movement directions and observes only the four adjacent cells, while previously observed cells remain in memory. An episode succeeds if the agent reaches the goal within 24 steps, requiring it to combine newly observed obstacles with information from earlier steps. In \textbf{WebShop}, we evaluate on 1,502 tasks sampled from a simulated web environment. 
We follow WebShop's native search-and-click interaction protocol, where the model receives the shopping instruction, action history, current page, and available actions at each step. {Each episode runs for at most 15 steps and receives a terminal reward ranging from 0 to 1, based on how well the purchased product matches the shopping instruction.}

\begin{table}[t]
    \centering
    \small
    \resizebox{\columnwidth}{!}{%
    \begin{tabular}{lrr}
        \toprule
        \multicolumn{3}{c}{\textbf{Grid}} \\
        \midrule
        Model & Success $\uparrow$ & Loop rate $\downarrow$ \\
        \midrule
        Qwen2.5-7B-Instruct  & 12.64\% & 81.2\% \\
        GPT-OSS-20B          & 54.57\% & 32.5\% \\
        Qwen2.5-32B-Instruct & 58.14\% & 41.6\% \\
        \midrule
        \multicolumn{3}{c}{\textbf{WebShop}} \\
        \midrule
        Model & Reward $\uparrow$ & Loop rate $\downarrow$ \\
        \midrule
        Qwen2.5-7B-Instruct  & 0.509 & 8.9\% \\
        GPT-OSS-20B          & 0.250 & 36.3\% \\
        Qwen2.5-32B-Instruct & 0.293 & 29.4\% \\
        \bottomrule
    \end{tabular}
    }
    \caption{
    Zero-shot performance (Success Rate on Grid and Reward on WebShop) of tested LLM agents. 
    Loop rate is the fraction of trajectories containing at least one local loop of actions (``\(A\!\rightarrow\!B\!\rightarrow\!A\!\rightarrow\!B\)''). 
    }
    \label{tab:baseline-agent-behavior}
\end{table}

\begin{figure*}[!t]
    \centering
    \includegraphics[width=\textwidth]{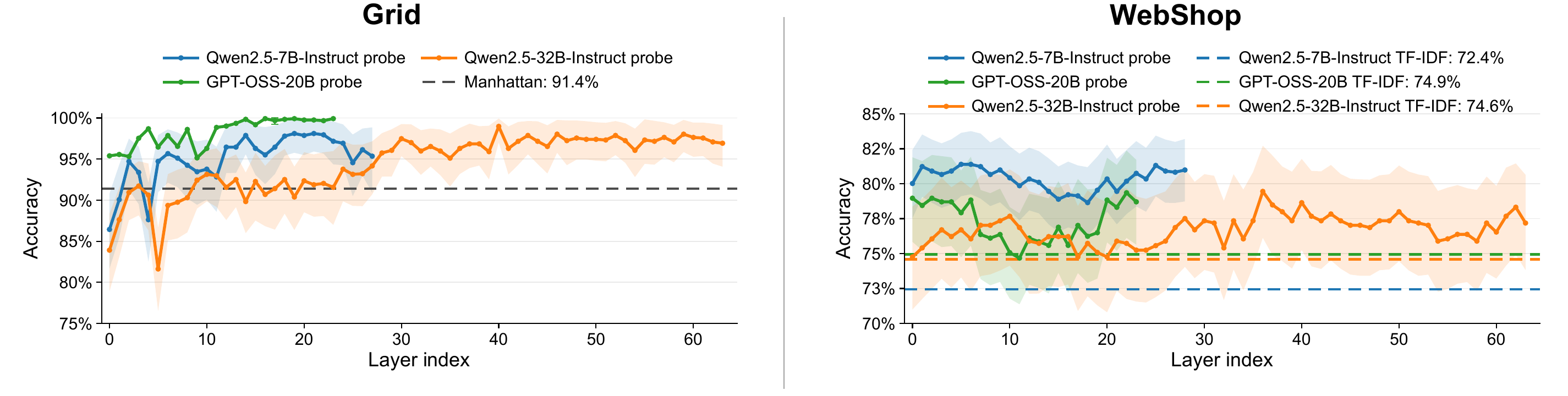}
    \caption{
    Layer-wise value readout in Grid and WebShop.
    (a) Grid held-out top-1 ranking accuracy, measuring whether the action with
    highest decoded value matches the gold best-value action; the dashed line shows
    the Manhattan baseline.
    (b) WebShop held-out pairwise ranking accuracy, measuring whether the readout
    correctly orders two candidate actions by empirical continuation value; dashed
    lines show the TF-IDF baselines.
    Shaded regions show 95\% cluster-bootstrap intervals for the learned readouts. 
    Results reported on a held-out validation set.
    }
    \label{fig:probe}
\end{figure*}

\paragraph{Model Performance} {Table~\ref{tab:baseline-agent-behavior} summarizes the behavior of the agents. In Grid, all models leave substantial room for improvement in reaching the goal. In WebShop, all models also achieve relatively low terminal reward. In addition, across both environments, we observed that agents sometimes exhibit repetitive behavior. We characterize the frequency of an agent showing repetitive behavior by its \emph{loop rate}, defined as the fraction of trajectories containing at least one local loop of actions following the pattern of ``\(A\!\rightarrow\!B\!\rightarrow\!A\!\rightarrow\!B\)''. The high loop rates of agents with varying LLM backends suggest that accumulated observations are not always translated into effective action choices. 

Surprisingly, in WebShop, the 32B Qwen model performs worse than its 7B variant. 
Similar comparisons were also found in previous agent evaluations, including Web Shopping in AgentBench and tool-use tasks in AgentBoard \citep{ICLR2024_e9df36b2,NEURIPS2024_877b4068}. One observation we made is that the 32B model often has a much longer navigation before making a purchase, leading to no purchase before the maximum time steps and hence zero terminal rewards. We further verify that the performance gap is not remedied by increasing the interaction budget by evaluating all models with a larger 30-step horizon (Appendix~\ref{app:stepsensitivity}). This extended exploration also leads to more frequent looping, where the agent repeatedly searches for or clicks on the same product.

{These performance limitations motivate our first mechanistic question: when an agent chooses an inferior action, is it because the agent does not model the value of potential actions properly?
We next formalize the quality of candidate actions in each environment and test whether this information is accurately represented in the agent's hidden states.}

\section{Value Is Represented but Does Not Determine Policy}
\label{sec:value_analysis}

The inconsistent behavior alone does not tell us whether an agent fails to represent its current environment, evaluate the future consequences of its actions inaccurately, or fails to use an already formed action evaluation. As a preliminary check in Grid, we find that the observed partial map can be linearly decoded from hidden states with very high accuracy
(Appendix~\ref{app:state}). This suggests that the inconsistent behavior is not simply
attributable to a general failure of state encoding \citep{li2023emergent, nanda-etal-2023-emergent},
which motivates us to examine representations more directly
related to planning. We therefore construct task-specific targets for prospective action value, fit linear readouts from hidden states, and compare the resulting value ordering with the model's policy.

\subsection{Prospective Action Value and Probing}
\label{sec:value_construction}

We use prospective action value to characterize the expected quality of candidate actions. 
Although we can provide a plausible construct for the action value quantification, this construct may not be the exact construct used by the agents. Therefore, instead of evaluating the absolute action value probing, we focus on \emph{action value contrast}, assuming that a better action is modeled with a higher value by the agent no matter if it adopts the same value construct as ours or not.
We summarize their construction below, with formal definitions and readout objectives provided in Appendix~\ref{app:value-targets}.

\paragraph{Grid.}
Because the environment dynamics are known, we evaluate each candidate move by its remaining shortest-path distance to the goal on the agent's current partial map. The map includes only information revealed so far, with unrevealed cells treated as free. We split the 1,400 maps into 1,200/100/100 corresponding to train/validation/test set. We then train a linear probe from the hidden state at the final prompt token to jointly predict the
values of the four movement actions.
{Out-of-bound actions are excluded because they do not correspond to valid successor states and therefore do not have a well-defined action value.}
We evaluate whether the readout ranks the highest-value action first. 

\paragraph{WebShop.}
Agents in WebShop take actions to either search for or click on products. While available click actions can be determined based on the present web page, the search actions are infeasible to predefine, since they are only generated by the agent at run time. Therefore, in this work, we only focus on probing the value of the click actions. Intuitively, the quality of an action depends on subsequent interactions, so we estimate it using sampled continuations.
We collect branch states along model-generated interaction prefixes and retain states with at least two distinct valid product-click candidates. We estimate each candidate's value from 40 sampled continuations under the model's policy. We extract mean candidate representations from their mentions in the prompt before action generation and train a linear classifier on their difference to predict which action in a pair has higher empirical continuation value. We evaluate the readout using pairwise ranking accuracy. See Appendix~\ref{app:webshop_candidates} for details.

\paragraph{Baselines.}
We compare the probes against two task-informed heuristic baselines. In Grid, Manhattan distance provides a simple but often effective
navigation strategy, although it ignores obstacles and may favor an incorrect move when a detour is necessary. In WebShop, TF-IDF ranks candidate actions by their lexical similarity to the shopping
instruction. This is a particularly strong reference for a shopping task, where matching requested attributes to products is already highly informative. Both baselines capture substantial task structure
without learned value representations, providing meaningful points of comparison for the readouts.

\subsection{Prospective Action Value Is Linearly Readable from Hidden States}
\label{sec:value_readout}

As shown in Figure~\ref{fig:probe}, prospective action value is linearly readable from intermediate hidden states in both environments and across model scales. 
In Grid, the readout becomes strong well before the final model output, indicating that future-outcome information is already organized in the residual stream. More importantly, the readout performs well at states where the Manhattan baseline is insufficient, such as when the agent must take a detour around a known obstacle. In these states, the Manhattan baseline gives a misleading ordering, whereas the learned readout still recovers the correct value ordering (see Appendix~\ref{app:grid-value-recovery} for details). This provides evidence that the model computes prospective value internally rather than simply following a geometric distance heuristic. Similar results extend to WebShop, where TF-IDF is already a strong baseline.

\subsection{Policy Errors Persist After the Correct Value Ordering Is Formed}
\label{sec:value_policy_dissociation}

A correct internal value ranking need not lead to the corresponding action. To quantify this gap, we jointly evaluate the greedy policy choice and the
selected value readout at the same decision states.
Table~\ref{tab:policy_probe_joint} cross-classifies their correctness for all three models in both environments.

\begin{table}[t]
\centering
\small
\setlength{\tabcolsep}{3pt}
\resizebox{\columnwidth}{!}{%
\begin{tabular}{@{}lrrrr@{}}
\toprule
 & \multicolumn{2}{c}{Policy correct}
 & \multicolumn{2}{c}{Policy wrong} \\
\cmidrule(lr){2-3}\cmidrule(lr){4-5}
Model & \shortstack{Probe\\correct}
      & \shortstack{Probe\\wrong}
      & \shortstack{Probe\\correct}
      & \shortstack{Probe\\wrong} \\
\midrule
\multicolumn{5}{@{}l}{\textit{Grid}} \\
Qwen2.5-7B-Instruct  & 39.6\% & 1.8\%  & \textbf{58.3\%} & 0.3\% \\
GPT-OSS-20B          & 62.2\% & 0.1\%  & \textbf{37.5\%} & 0.2\% \\
Qwen2.5-32B-Instruct & 64.0\% & 0.2\%  & \textbf{34.2\%} & 1.6\% \\
\midrule
\multicolumn{5}{@{}l}{\textit{WebShop}} \\
Qwen2.5-7B-Instruct  & 40.4\% & 8.3\%  & \textbf{43.9\%} & 7.4\% \\
GPT-OSS-20B          & 42.5\% & 12.7\% & \textbf{39.8\%} & 5.0\% \\
Qwen2.5-32B-Instruct & 40.4\% & 9.5\%  & \textbf{39.8\%} & 10.3\% \\
\bottomrule
\end{tabular}%
}
\caption{
{Joint correctness of policy decisions and value readouts. Each entry reports the percentage of decision states falling into one of four categories:
whether the policy choice is correct and whether the value probe correctly identifies the better action.
Bold entries highlight states where the policy chooses the worse action while the probe correctly identifies the better one.}
}
\label{tab:policy_probe_joint}
\end{table}

The mismatch is strongly asymmetric. In Grid, the policy is wrong while the readout is correct on 34.2\%--58.3\% of evaluated states, compared with only 0.1\%--1.8\% in the opposite case. In WebShop,
the corresponding fractions are 39.8\%--43.9\% versus
8.3\%--12.7\%. Thus, value information can be linearly recoverable from hidden states without consistently determining the model's selected action.

 These results show that poor decisions need not arise because the model failed to represent the relevant action values. They may instead reflect a failure to use this value information to guide the policy. {This suggests that value information is not reliably translated into action selection, a phenomenon we call \emph{value-policy dissociation}. We therefore investigate whether interventions on value-related hidden representations can make this information more effective in guiding the policy toward higher-value actions.}
 

\section{From Value Readout to Policy Control}
\label{sec:policy_control}

\begin{figure*}[t]
\centering
\includegraphics[width=\linewidth]{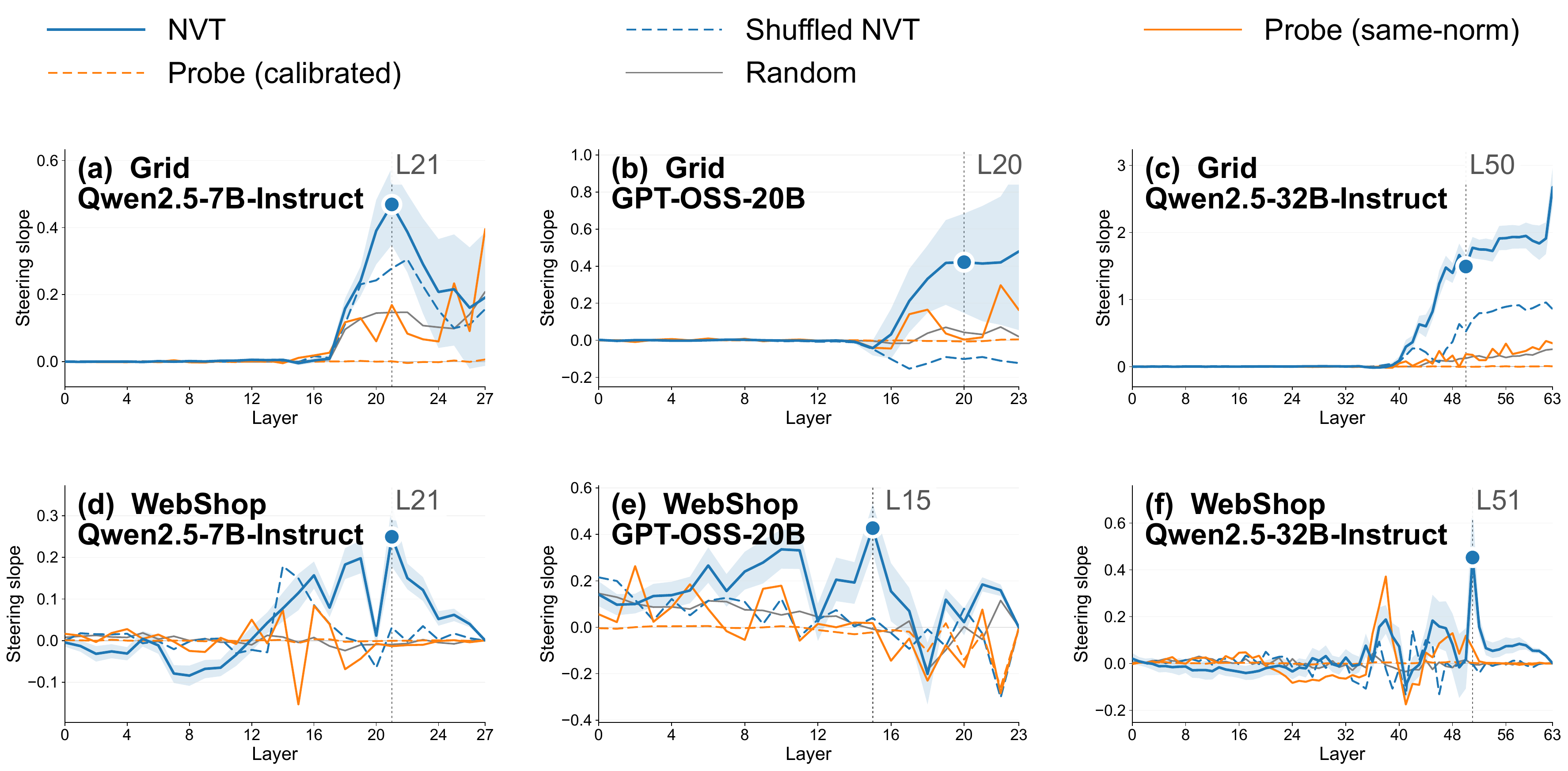}
\caption{
Layer-wise steering slope for Natural Value Transport (NVT) and control interventions across Grid and WebShop. Probe (calibrated) uses the scale implied by the value readout; Probe (same-norm), Shuffled NVT, and Random
match the magnitude of the NVT perturbation while changing its direction. Positive slope indicates increased policy preference for the action favored by the value readout. NVT identifies a more stable late-layer policy-responsive region across tasks. Results are reported on a held-out validation set.
}
\label{fig:policy_control_layers}
\end{figure*}

In this section, we explore intervention approaches to enable the value readout to shift the agent's action selection.
We compare probe-direction steering with \emph{Natural Value Transport (NVT)}, which uses hidden-state changes naturally associated with differences in prospective value.

\subsection{Intervention Design}
\label{sec:natural_value_map}

{For a decision state $s$, we consider a pair
$(a^+,a^-)$, where $a^+$ is favored by the value
readout and $a^-$ is the alternative. 
All methods construct an edit
$\boldsymbol{d}_\ell\in\mathbb{R}^{D_\ell}$
scaled by an intervention strength $\alpha$.
In Grid, the edit is added to the final-token
activation; in WebShop, it is applied
symmetrically to the two candidate spans,
changing their representation difference by
$\alpha\boldsymbol{d}_\ell$.
The methods differ in how
$\boldsymbol{d}_\ell$ is constructed.}


\paragraph{Steering along Value Readout Directions.}
{A natural baseline is to steer along directions
identified by the fitted value probe, following
prior inference-time interventions
\citep{NEURIPS2023_81b83900,yan2026interpreting}.
Let $\delta\boldsymbol{v}\in\mathbb{R}^{k}$ denote the
desired change in the decoded value coordinates,
where $k=4$ in Grid and $k=1$ in WebShop.
\emph{Probe (calibrated)} constructs the edit
\(\boldsymbol{d}_\ell^{\mathrm{probe\text{-}cal}} = P_\ell \delta \boldsymbol{v} \in \mathbb{R}^{D_\ell}\),
where $P_\ell\in\mathbb{R}^{D_\ell\times k}$ is a
task-specific probe-aligned calibration map.
In Grid, it is obtained from the pseudoinverse
of the four-output linear readout; in WebShop,
it projects the NVT edit onto the pairwise
probe direction.}

\paragraph{Natural Value Transport.}
{An alternative is to learn how hidden states vary when the represented values of available actions differ. To this end, we propose \emph{Natural Value Transport (NVT)}, which explicitly learns a projection from the value changes to the corresponding hidden-state changes; the changes finally form a dynamic steering vector $\boldsymbol{d}_\ell$.
Specifically, each training sample $i$ corresponds to a pair of comparisons $(a_i^+,a_i^-)$ counterfactually constructed for the training purpose. Let
$\Delta\boldsymbol{v}_i\in\mathbb{R}^k$ and
$\Delta\boldsymbol{h}_{\ell,i}\in\mathbb{R}^{D_\ell}$
be the differences in value and layer-$\ell$ activations from the same comparison. 
In Grid, we compare matched observations with a cell marked blocked or free, using differences in the four decoded action distances ($k=4$).
In WebShop, two actions $(a_i^+,a_i^-)$ at state $s_i$ give
$\Delta\boldsymbol{v}_i=\widehat Q^\pi(s_i,a_i^+)
-\widehat Q^\pi(s_i,a_i^-)$ and
$\Delta\boldsymbol{h}_{\ell,i}
=\boldsymbol{h}_\ell(s_i,a_i^+)
-\boldsymbol{h}_\ell(s_i,a_i^-)$,
where $\widehat Q^\pi$ is the estimated continuation value and $\boldsymbol{h}_\ell(s,a)$ is the candidate action's hidden representation.
Each WebShop comparison therefore provides a single scalar value difference ($k=1$). We fit $B_\ell\in\mathbb{R}^{D_\ell\times k}$ by
\begin{equation}
B_\ell=\operatorname*{arg\,min}_{B}\sum_i w_i
\|\Delta\boldsymbol{h}_{\ell,i}-B\Delta\boldsymbol{v}_i\|_2^2
+\lambda_B\|B\|_F^2
\end{equation}
where $w_i$ weights training comparisons and $\lambda_B$ is the ridge regularization coefficient. At inference time, given a value change
$\delta\boldsymbol{v}\in\mathbb{R}^{k}$,
NVT defines the edit $\boldsymbol{d}_\ell^{\mathrm{NVT}}=B_\ell\delta\boldsymbol{v}\in\mathbb{R}^{D_\ell}$.
Thus, NVT applies the value-projected activation changes rather than the probe direction for steering.

\paragraph{Controls.}
To separate steering direction from perturbation magnitude, we also compare NVT with \emph{Probe (same-norm)}, a variant of Probe (calibrated) that matches the norm of NVT, i.e., \(\boldsymbol{d}_\ell^{\mathrm{probe\text{-}norm}}=\|\boldsymbol{d}_\ell^{\mathrm{NVT}}\|_2\widetilde{\boldsymbol{u}}_\ell.\), where $\widetilde{\boldsymbol{u}}_\ell\in\mathbb{R}^{D_\ell}$ denotes the unit probe-aligned direction oriented toward $a^+$.
The second control, \emph{Shuffled NVT}, shuffles and breaks the value--activation pair when training NVT while matching the norm.
If the learned correspondence is important for
policy control, disrupting it should weaken the
systematic policy response.
Finally, \emph{Random} uses an unrelated direction with
the same norm, testing for generic perturbation
effects.}

\begin{table*}[t]
    \centering
    \small
    \begin{tabular}{llrccc}
        \toprule
        Task & Model & $n$
        & Outcome $\uparrow$
        & Loop $\downarrow$
        & Flip rate \\
        \midrule
        Grid
        & Qwen2.5-7B-Instruct
        & 100
        & 0.140 $\rightarrow$ 0.300
        & 85.0\% $\rightarrow$ 64.0\%
        & 24.2\% \\
        Grid
        & GPT-OSS-20B
        & 100
        & 0.580 $\rightarrow$ 0.640
        & 43.0\% $\rightarrow$ 38.0\%
        & 16.9\% \\
        Grid
        & Qwen2.5-32B-Instruct
        & 100
        & 0.590 $\rightarrow$ 0.680
        & 41.0\% $\rightarrow$ 32.0\%
        & 37.3\% \\
        WebShop
        & Qwen2.5-7B-Instruct
        & 671
        & 0.473 $\rightarrow$ 0.501
        & 7.2\% $\rightarrow$ 6.1\%
        & 42.3\% \\
        WebShop
        & GPT-OSS-20B
        & 503
        & 0.417 $\rightarrow$ 0.458
        & 15.5\% $\rightarrow$ 14.9\%
        & 16.5\% \\
        WebShop
        & Qwen2.5-32B-Instruct
        & 454
        & 0.358 $\rightarrow$ 0.452
        & 2.0\% $\rightarrow$ 1.8\%
        & 55.1\% \\
        \bottomrule
    \end{tabular}
    \caption{
    Behavioral effects of value steering.
    Paired entries report baseline $\rightarrow$ steered performance evaluated over $n$ episodes. Outcome denotes success rate for Grid and mean reward for WebShop. Flip rate is the fraction of evaluated decisions whose selected action changes under the intervention. 
    }
    \label{tab:steering_behavior}
\end{table*}

\paragraph{Measuring Policy Response}

{An effective intervention should produce a systematic
dose response toward $a^+$.
We summarize this response by \emph{steering slope},
the fitted change in
$\log\pi_{\ell,\alpha}(a^+\mid s)
-\log\pi_{\ell,\alpha}(a^-\mid s)$
per unit increase in $\alpha$.
Positive slopes indicate increasing preference for the readout-favored action. Full construction details are provided in Appendix~\ref{app:Mathmaticaldetails}.}

\subsection{Natural Value Transport Reveals a Policy-Relevant Region}
\label{sec:direct_probe}

As shown in Figure~\ref{fig:policy_control_layers}, \emph{Probe (calibrated)} yields little policy response across layers: changing a probe-readable value is not, by itself, a reliable way to change the policy.
\emph{Probe (same-norm)} produces positive slopes at some layers, suggesting overlap between readout directions and policy-sensitive subspaces. These effects, however, are localized and do not form a consistent pattern across tasks.

In contrast, \emph{NVT} exhibits a more coherent late-layer response. For Qwen2.5-7B-Instruct, the Grid and WebShop results peak in similar late-layer regions; GPT-OSS-20B and Qwen2.5-32B-Instruct also show
late-layer policy-sensitive regimes. Neither \emph{Shuffled NVT} nor \emph{Random} reproduces this pattern consistently. The difference therefore cannot be explained by perturbation magnitude alone: the direction of the value-associated hidden-state change matters for steering the policy.

Together, these results separate \emph{value readout} from \emph{value control}. A probe identifies directions along which value is decodable and sometimes causally influential, whereas NVT more consistently identifies late-layer edits that shift action preference toward higher-value options.

\section{Steering Agent Decisions with Prospective Value}
\label{sec:steering}

Section~\ref{sec:direct_probe} identifies value-associated edits that can change action preference. We now test whether applying such edits during interaction improves realized behavior.
We combine a trained value readout (\emph{sensor}), which identifies a preferred action, with NVT (\emph{actuator}), which supplies the hidden-state edit. The two components can operate at different layers.

\subsection{Held-Out Value-Guided Control}
\label{sec:heldout_controller}

We compare unsteered and steered rollouts on matched evaluation instances. In Grid, both runs begin at the test map (a held-out set of 100 maps)'s initial state, and steering is reconsidered after each transition until termination or the step limit. 
In WebShop, we steer decisions involving
multiple clickable products along test trajectories containing at least one such decision. WebShop uses five task-disjoint folds, fitting the readout and NVT map on four folds and calibrating only on training data.
Model-specific configurations appear in
Appendix~\ref{app:steering_settings}.

\subsection{Value-Guided Steering Improves Behavior}
\label{sec:heldout_steering}

\paragraph{Outcome gains.}
Table~\ref{tab:steering_behavior} summarizes the behavioral effects of value-guided steering. Across both environments and all three models, steering improves task outcomes on the evaluated instances.

In Grid, steering improves success and reduces looping across all three models, with the largest improvement for Qwen2.5-7B-Instruct. The other two models improve on both metrics, though the gains are smaller. The observed flips further show that the controller can alter action selection during interaction, rather than merely changing the log-preference margin.

In WebShop, steering improves realized reward across all three models, while the reductions in looping are more modest. One possible limitation is the imperfect value readout: the WebShop probe achieves around 80\% pairwise ranking accuracy and may occasionally favor an action whose value is actually lower. When steering is applied repeatedly over more complex trajectories, such errors may offset some of the gains from other decisions.

\begin{table}[t]
    \centering
    \small
    \begin{tabular}{llc}
        \toprule
        Setting & Model & Invalid rate $\downarrow$ \\
        \midrule
        Grid    & Qwen2.5-7B-Instruct  & 0.0\% $\rightarrow$ 0.17\% \\
        Grid    & GPT-OSS-20B & 0.0\% $\rightarrow$ 0.0\%  \\
        Grid    & Qwen2.5-32B-Instruct & 0.0\% $\rightarrow$ 0.0\%  \\
        WebShop & Qwen2.5-7B-Instruct  & 0.0\% $\rightarrow$ 1.0\%  \\
        WebShop & GPT-OSS-20B & 0.1\% $\rightarrow$ 0.4\%  \\
        WebShop & Qwen2.5-32B-Instruct & 9.7\% $\rightarrow$ 6.2\%  \\
        \bottomrule
    \end{tabular}
    \caption{
    Invalid-action rates under value steering.
    Entries report baseline $\rightarrow$ steered rates.
    }
    \label{tab:steering_invalid}
\end{table}


\paragraph{Robustness check.}
{One possible concern is that altering internal computations may improve some actions while harming
others \citep{aghzal2026agentmi}.} Therefore, we also examine whether steering affects the frequency of invalid actions, as shown in Table~\ref{tab:steering_invalid}.
In Grid, the value readout is trained only on valid actions. We therefore replace proposed moves into known obstacles or out of bounds with the highest-scored valid action before execution. The remaining invalid Grid actions are outputs that cannot be parsed into valid directions.

In WebShop, invalid actions include formatting
violations and well-formed but unexecutable
actions, such as clicking an unavailable product. Overall, steering improves task outcomes without consistently reducing invalid actions.

\section{Related Work}
\label{sec:related_work}

\paragraph{World representations and planning in language models.}
Mechanistic interpretability work has shown that sequence models develop structured internal representations of their environments, including board states in Othello and chess \citep{li2023emergent,nanda-etal-2023-emergent,karvonen2024emergent,NEURIPS2024_37d9f191}, spatial and temporal structure in pretrained LMs \citep{ICLR2024_0a605985}, and spatial layouts, agent states, and multi-step plans in grid-world agents \citep{martorell2025text,arghal2026a}. Controlled navigation and path-planning benchmarks also study LLMs' spatial and long-horizon planning abilities \citep{aghzal2024can,aghzal2024look}. Other work studies prospective information such as future tokens, decisions, and response-level plans \citep{pal-etal-2023-future,men-etal-2024-unlocking,dong2025emergent,hanna2026latent}. These studies mainly examine representations of world states or future behavior; whether LLM agents evaluate the downstream value of \emph{alternative} actions during sequential decision making remains less understood.

\paragraph{Value estimation and sequential decision making.}
Value functions are central to reinforcement learning, where state and action values summarize expected future return and support policy improvement \citep{sutton1991dyna,ha2018world,schrittwieser2020mastering}. For LLM agents, planning methods use explicit world-model simulation and task rewards \citep{hao-etal-2023-reasoning,gu2024your}, learned Q-value or reward models guide action selection and test-time search \citep{DBLP:conf/aaai/ZhaiYXFYDW25,xia-etal-2025-agentrm}, and expected reward can be estimated directly from an actor LM's hidden states \citep{choi2026your}. These studies treat value as a quantity to be learned or added for training and search, rather than as an internal computation to be studied mechanistically in an otherwise frozen LLM agent.

\paragraph{Mechanistic analysis and intervention in agent decisions.}
Probes identify encoded information in neural representations \citep{alain2016understanding,belinkov-2022-probing}, while causal mediation, activation patching, and representation steering test how internal computations influence model outputs and enable inference-time interventions \citep{elazar-etal-2021-amnesic,vig2020causal,NEURIPS2022_6f1d43d5,zou2023representation}. Recent work shows that highly decodable features need not be the main drivers of generation \citep{yan2026interpreting}, and that multi-step, environment-dependent trajectories pose new challenges for mechanistic analysis of agents \citep{aghzal2026agentmi}. Closely related, \citet{okcu2026beliefactiongap} combine state probes, inverse-reinforcement-learning-based value estimation, and emitted actions to study belief--action gaps. We further study prospective value at the level of candidate actions in both partially observed navigation and WebShop, and use causal interventions to examine how value-related representations relate to subsequent policy decisions.
\section{Conclusion}
\label{sec:conclusion}

We find that LLM agents can encode the relative downstream value of candidate actions without reliably using that information to determine their actions. Across Grid and WebShop, value information is linearly decodable from intermediate hidden states, while NVT interventions more consistently influence action preferences at later layers. Leveraging these value-associated interventions improves behavior. These results identify value use, rather than value representation alone, as a distinct locus of agent failure and provide a mechanistic target for diagnosing decisions that cannot be understood from behavior alone.

\section*{Limitations}
\label{sec:limitations}
Our study has several limitations. First, we study three models from two families across two sequential decision environments, Grid and WebShop. Although these settings provide complementary forms of prospective action value, it remains unclear whether the same value--policy dissociation and policy-facing structure generalize to other model families, agent architectures, or substantially different decision domains. Second, our notion of prospective value is necessarily task dependent: Grid uses an environment-derived shortest-path target, whereas WebShop uses finite-sample estimates of on-policy continuation return. These operationalizations capture useful downstream consequences but do not exhaust possible notions of value, and the WebShop estimates are inherently sampling-dependent and policy-dependent. Third, in WebShop, our method considers only click actions and may not readily generalize to search actions. Fourth, our mechanistic analysis focuses on linear readouts and a linear value-to-hidden-state transport operator. Their causal effects support a separation between value representation and policy control, but they do not constitute a complete account of how value information is computed or propagated through the model. Finally, our steering procedure requires white-box access to internal activations and task-specific data for fitting the readout and transport operator, which limits its direct applicability to closed-source models and settings where such supervision is unavailable.

\section*{Acknowledgments}
M. A. and Z. Y. were sponsored by the National Science Foundation (NSF) Grant No. 2418580 and 2540626 and the Foresight Institute. 



\bibliography{custom}

\appendix
\newpage

\section{Prospective Action Value Targets and Readout Objectives}
\label{app:value-targets}

\subsection{Grid.}
Let $\mathcal{M}(s)$ denote the agent's current partial map at decision state $s$, containing all cells revealed so far, with unrevealed cells treated as free. For a candidate action $a$,
let $s_a$ denote the resulting state. We define
\[
V(s,a) =-\, d_{\mathcal{M}(s)}(s_a,g),
\]
where $g$ is the goal and $d_{\mathcal{M}(s)}(\cdot,\cdot)$ is
the shortest-path distance on the current partial map.
This is a distance-based value of the successor state, rather than a policy-conditioned estimate of the return. It depends only on information available to the agent at the current decision state.

At each layer $\ell$, we extract the hidden state at the final prompt token, denoted $h_\ell(s)$, and fit a linear readout
\[
\widehat{\mathbf{V}}_\ell(s)
= W_\ell h_\ell(s) + b_\ell,
\qquad
\widehat{\mathbf{V}}_\ell(s)
\in \mathbb{R}^{|\mathcal{A}(s)|}.
\]
In Grid, $|\mathcal{A}(s)|=4$, and each component corresponds to one movement action. We evaluate the readout by whether it ranks the gold best-value action first. The Manhattan baseline instead ranks actions by their Manhattan distance to the goal, without accounting for obstacles.

\subsection{WebShop.}
\label{app:webshop_candidates}
For a branch state $s$ and candidate action $a$, we define the policy-conditioned continuation value
\[
Q^\pi(s,a)
= \mathbb{E}\!\left[
R \mid s_t=s,\; a_t=a,\;
\pi \text{ thereafter}
\right],
\]
where $R$ is the final task reward and $\pi$ is the model's
continuation policy. We estimate this quantity using $C=40$
sampled continuations:
\[
\widehat{Q}^{\pi}(s,a)
= \frac{1}{C}\sum_{c=1}^{C} R^{(c)}(s,a).
\]

\paragraph{Candidate selection in WebShop.}
Although WebShop permits free-form \texttt{search[query]} actions, we do not enumerate the full textual action space. Instead, we consider legal product-click actions on the current page. We follow model-generated greedy interaction prefixes and identify states that expose at least two legal product IDs and for which
the model's greedy action is a valid product click.

At such a state, we draw 48 action samples from the same model (temperature = $0.7$, top-$p$ = $0.9$), retaining only valid product clicks. We require at least eight valid product samples, at least one sampled action matching the greedy action, and at least one distinct alternative. We then retain the greedy action and up to two most frequent distinct sampled alternatives.

Each retained candidate is evaluated through sampled policy continuations, and only pairs with different empirical continuation values are used for the pairwise readout.

\paragraph{Pre-decision candidate representations.}
All WebShop candidate activations are extracted from the
input at the current decision state, \emph{before the model generates its next action}. When a product identifier appears multiple times in
the prompt, we use its occurrence in the
\texttt{Allowed click targets} list. For a candidate $a$, $h_\ell(s,a)$ is the mean activation over the token span identifying that candidate \emph{in the decision prompt}, not in a generated response or an executed action.
For a pair $(a_i,a_j)$, we form
\[
\Delta h_{\ell,ij}
= h_\ell(s,a_i)-h_\ell(s,a_j)
\]
and train a linear classifier to predict whether
\[
\widehat{Q}^{\pi}(s,a_i)
>
\widehat{Q}^{\pi}(s,a_j).
\]
We evaluate the readout using pairwise action-ranking accuracy. The TF-IDF baseline ranks candidate actions by lexical similarity to the shopping instruction.

\subsection{Justification of Value Target Construction Across Environments}
The value targets in Grid and WebShop are constructed using different procedures due to the distinct characteristics of the two environments. These choices reflect the difference between a fully specified planning environment and an open-ended interactive environment, rather than representing different notions of value.

In Grid, the environment dynamics are deterministic and the complete state space is available. Therefore, the optimal value of an action can be directly computed from the remaining distance to the goal. We define the action value based on the number of steps required to reach the target state under the optimal path, which provides a precise and noise-free estimate of the downstream utility of each action. This formulation is consistent with classical shortest-path planning, where fewer remaining steps correspond to higher expected return.

In contrast, WebShop involves a much more complex setting. Unlike Grid, there is no predefined transition model or oracle optimal trajectory available for computing exact action values. Instead of annotating the value manually, we estimate value from the agent's own trajectories by sampling rollouts and measuring the resulting task-level outcomes. This empirical return-based estimation captures the utility that an agent can reasonably infer from its interaction history and is aligned with the standard reinforcement learning setting, where values are learned from sampled experience. Furthermore, rather than training a regression probe to predict the absolute value of an action, we formulate the probe as a ranking classifier since it is unrealistic to assume that a model can accurately estimate the exact action value in such a complex environment.

Although the two environments use different value estimation procedures, both definitions serve the same purpose: quantifying the expected quality of an action with respect to future outcomes. Using exact planning-based values in Grid and trajectory-based empirical values in WebShop allows us to obtain the most faithful estimates of action value available in each environment while avoiding introducing unnecessary assumptions about hidden dynamics in the interactive setting.

We deliberately avoid delegating action selection to an external Q-network or retraining the policy through reinforcement learning, as our goal is to investigate how internal value representations can causally influence the model's own action selection through direct hidden-state interventions. We apply steering to test the models' performance gains in Section~\ref{sec:heldout_steering} instead of directly teacher-forcing the best readout action for the same reason.

\section{State and Global Value-Map Readout in Grid}

\subsection{State representation}
\label{app:state}
We first test whether the prompt-last hidden state contains a readable representation of the observed Grid map. At each layer, a linear readout predicts one of three labels for each of the 25 cells,
\[
    h_\ell(s)
    \longmapsto
    \widehat{\mathbf M}_\ell(s)\in\{\mathrm F,\mathrm O,\mathrm U\}^{25},
\]
where \(\mathrm F\), \(\mathrm O\), and \(\mathrm U\) denote known free, known obstacle, and unrevealed cells. All 25 cells participate in the classification.
Table~\ref{tab:grid_state_map} reports cell accuracy and exact-map accuracy, which requires all 25 labels to be correct. The probes recover the observed map with high accuracy; the positional-majority baseline reaches about 61\% cell accuracy but 0\% exact-map accuracy.

\begin{table}[t]
    \centering
    \caption{Readout of the 25-cell F/O/U state map. Probe layers are selected by validation macro-F1. The majority baseline assigns each position its most frequent training label. Exact-map accuracy requires all 25 cells to be correct.}
    \label{tab:grid_state_map}
    \footnotesize
    \setlength{\tabcolsep}{2pt}
    \begin{tabular}{@{}lrrrr@{}}
        \toprule
        & \multicolumn{2}{c}{Probe} & \multicolumn{2}{c}{Majority Baseline} \\
        \cmidrule(lr){2-3}\cmidrule(l){4-5}
        Model & Cell & Exact map & Cell & Exact map \\
        \midrule
        Qwen2.5-7B-Instruct  & 98.20 & 79.73 & 60.69 & 0.00 \\
        Qwen2.5-32B-Instruct & 97.89 & 79.02 & 60.69 & 0.00 \\
        GPT-OSS-20B & 96.28 & 72.18 & 61.46 & 0.00 \\
        \bottomrule
    \end{tabular}
\end{table}

\subsection{Global distance}
\label{app:grid-value-recovery}
We next test whether the hidden state contains a more global value representation than the four action values considered in the main text. At each layer, we fit a linear readout from the prompt-last hidden state to the
optimistic remaining distance for every cell in the \(5\times5\) grid,
\[
    h_\ell(s)
    \longmapsto
    \widehat{\mathbf V}_\ell(s)\in\mathbb{R}^{25}.
\]
Equivalently, the readout contains one scalar linear prediction for each grid position. The target for each position is its shortest-path distance to the
goal under the currently observed map: known obstacles are treated as blocked, while unrevealed cells remain traversable. Cells corresponding to known obstacles or otherwise undefined targets are masked from the regression loss and evaluation.

Figure~\ref{fig:global_vmap_layers} shows the resulting layer-wise readout error for all three models. The global prediction becomes substantially more accurate in intermediate and later layers. The same pattern holds when
evaluation is restricted to \emph{detour cells}, where known obstacles make the shortest-path distance differ from Manhattan distance. The probe strongly
outperforms the Manhattan baseline on these cells, indicating that the decoded representation contains obstacle-aware spatial structure rather than only the
goal location or a geometric distance shortcut. Since the target distances are integers, the lower row also reports MAE after rounding each predicted distance to the nearest integer, using the same evaluation cells.

\begin{figure}[t]
    \centering
    \includegraphics[width=\columnwidth]{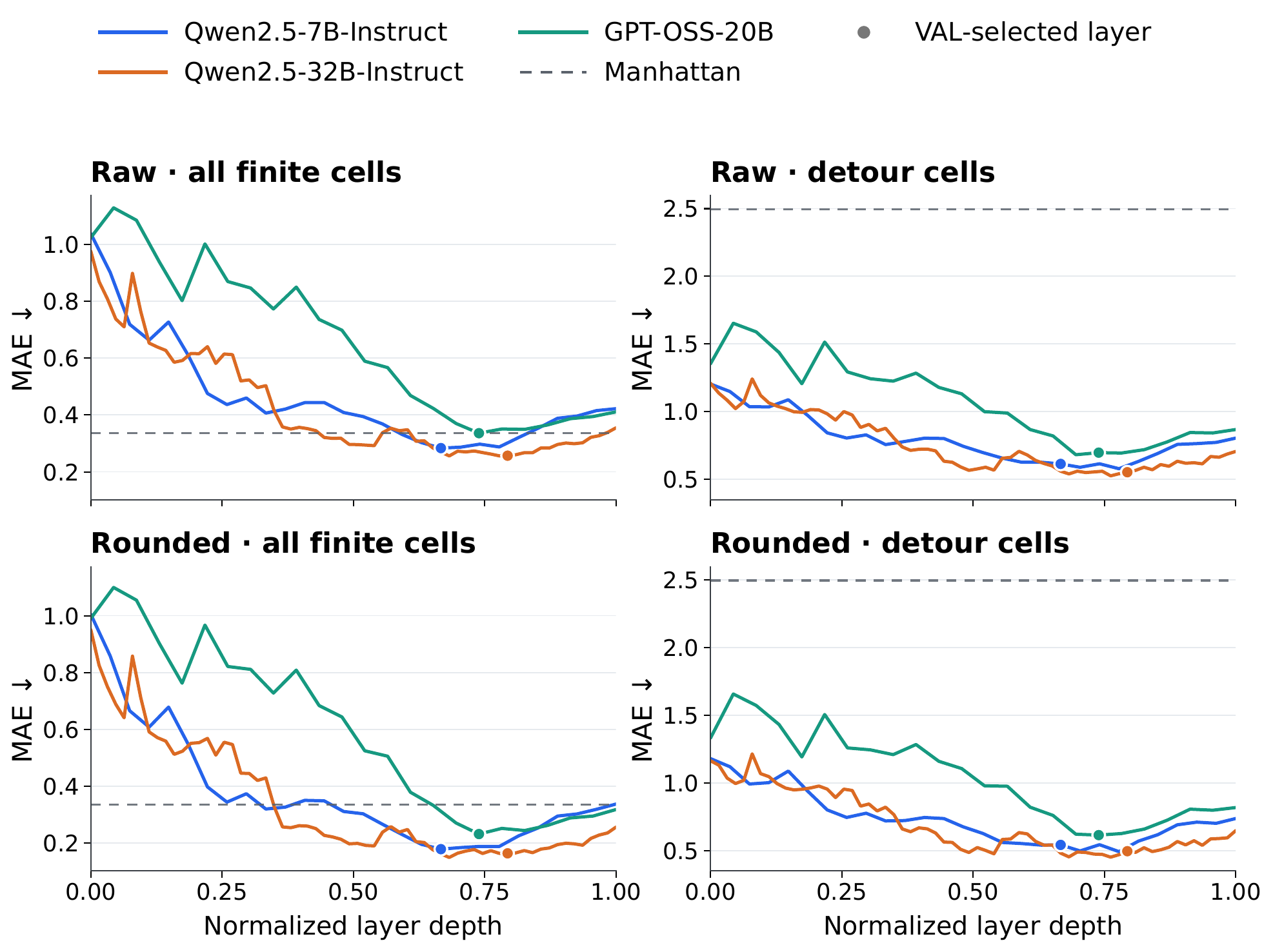}
    \caption{
    Layer-wise linear readout of the full Grid value map. Left: MAE over all
    finite cells in the \(5\times5\) optimistic distance-to-goal field. Right:
    MAE on detour cells, where known obstacles make the shortest-path distance
    differ from Manhattan distance. Top: continuous predictions. Bottom:
    predictions rounded to the nearest integer before computing MAE. Dashed
    and dotted lines denote the Manhattan baselines for the shared Qwen states
    and the GPT-OSS native states, respectively. Markers indicate layers
    selected by validation detour-cell MAE.
    }
    \label{fig:global_vmap_layers}
\end{figure}

\begin{figure}[!htbp]
    \centering
    \includegraphics[width=0.88\linewidth]{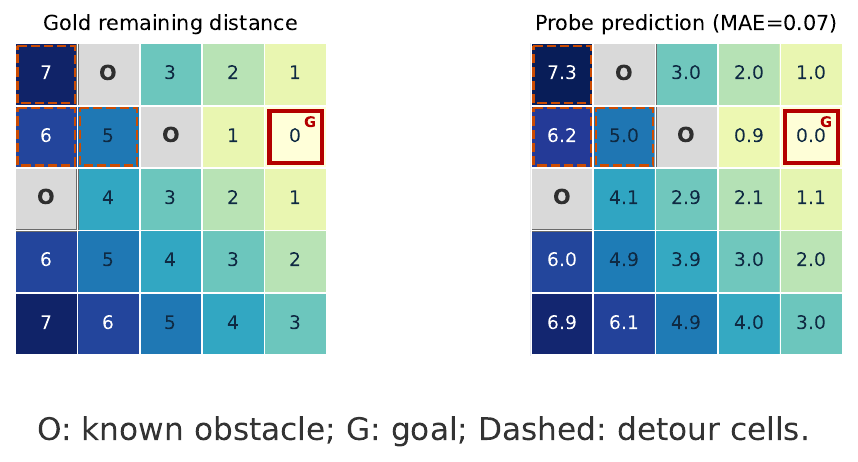}
    \caption{
    Example global value-map readout on a held-out Grid state.
    Left: gold optimistic remaining distance to the goal.
    Right: prediction from the linear readout.
    Known obstacles are marked by \texttt{O}, the goal by \texttt{G}, and
    dashed cells indicate locations whose shortest-path distance differs from
    Manhattan distance because of the observed obstacles.
    }
    \label{fig:global_vmap_example}
\end{figure}

\begin{table*}[t]
\centering
\caption{
WebShop performance under different maximum interaction budgets. All models are evaluated on the same 1,502 shared tasks. Reward is averaged over tasks, purchase rate denotes the fraction of episodes containing a
\texttt{click[Buy Now]} action, and loop rate follows the trajectory-level loop definition used in the main experiments.
}
\label{tab:webshop_step_sensitivity}
\begin{tabular}{llcccc}
\toprule
Model & Max Steps & Reward & Purchase Rate & Mean Steps & Loop Rate \\
\midrule
Qwen2.5-7B-Instruct & 15 & 0.509 & 76.0\% & 6.63 & 8.9\% \\
Qwen2.5-7B-Instruct & 30 & 0.517 & 78.0\% & 9.80 & 9.5\% \\
\midrule
GPT-OSS-20B & 15 & 0.250 & 41.0\% & 10.89 & 36.3\% \\
GPT-OSS-20B & 30 & 0.284 & 46.9\% & 19.43 & 53.3\% \\
\midrule
Qwen2.5-32B-Instruct & 15 & 0.293 & 51.5\% & 7.01 & 29.4\% \\
Qwen2.5-32B-Instruct & 30 & 0.297 & 52.8\% & 11.28 & 30.0\% \\
\bottomrule
\end{tabular}
\end{table*}

Figure~\ref{fig:global_vmap_example} shows one held-out state with several known obstacles. A single prompt-last hidden state is used to predict the remaining distance at every non-obstacle position in the grid. The decoded
map closely matches the gold distance field, including cells whose values are changed by obstacle-induced detours. This provides a more direct view of the
representation underlying the action-value results in the main text: value information is not limited to the immediately available actions, but is organized as a global prospective value field over the observed environment.

\section{Effect of Interaction Budget on WebShop Performance}
\label{app:stepsensitivity}

Our main WebShop evaluation uses a maximum interaction budget of 15 steps (the default setting of WebShop). To examine whether the observed performance differences are primarily caused by insufficient interaction budgets, we additionally evaluate all models with a larger budget of 30 steps. We perform this analysis on the same set of 1,502 shared tasks used in our behavioral analyses, starting each episode from the initial environment state and using the same greedy decoding protocol.

Table~\ref{tab:webshop_step_sensitivity} summarizes the results. Increasing the step budget has limited impact on the relative ordering of models. Qwen2.5-7B-Instruct remains substantially stronger than the larger models under both budgets. GPT-OSS-20B obtains a modest improvement with additional steps, suggesting that some failures are related to insufficient interaction length; however, its performance remains substantially below the 7B model. Qwen2.5-32B-Instruct shows little change when increasing the interaction budget, indicating that its lower performance is not primarily explained by prematurely exhausting the available steps.

These results suggest that increasing the allowed interaction horizon alone does not resolve the decision-making gap between models. While GPT-OSS-20B benefits from additional steps, the improvement is insufficient to match the performance of Qwen2.5-7B-Instruct. The results indicate that the observed differences are not primarily explained by the external interaction budget.

\section{Mathematical Details of Value-Directed Interventions}
\label{app:steering_math}

\subsection{Mathematical details}
\label{app:Mathmaticaldetails}

We detail the constructions compared in
Sections~\ref{sec:natural_value_map}.
All hidden-state edits are expressed in $D_\ell$-dimensional coordinates at layer $\ell$. Training differences are denoted by $\Delta\boldsymbol{v}_i$ and
$\Delta\boldsymbol{h}_{\ell,i}$; the intervention-time command is $\delta\boldsymbol{v}$. Bold symbols denote vectors; $B_\ell$ and $W_\ell$ denote
matrices. Scalar WebShop value differences are identified with one-dimensional vectors in the common NVT objective.
The value coordinates are task-specific: Grid uses four decoded remaining-distance scores (smaller is better), while WebShop uses a scalar continuation-reward difference (larger is better). Grid's distance scores are the negatives of the value convention in
Appendix~\ref{app:value-targets}; changing the sign of both the input coordinates and the fitted transport yields the same hidden-state edit.

\paragraph{NVT objective and training differences.}
For paired training observations indexed by $i$, define
$\Delta\boldsymbol{h}_{\ell,i}\in\mathbb{R}^{D_\ell}$ and
$\Delta\boldsymbol{v}_i\in\mathbb{R}^{k}$, with $k=4$ in Grid and $k=1$ in
WebShop. We fit
\begin{equation}
 \begin{aligned}
 B_\ell=\arg\min_B\;&\sum_i w_i
  \|\Delta\boldsymbol{h}_{\ell,i}-B\Delta\boldsymbol{v}_i\|_2^2\\
  &+\lambda_B\|B\|_F^2
 \end{aligned}
 \label{eq:app_nvt_fit}
\end{equation}
The weight $w_i$ balances training comparisons and $\lambda_B>0$ regularizes the map. The learned map is fitted on training examples only.

In \emph{Grid}, each pair comes from a newly observed obstacle at a single grid cell. Let $s_i^O$ be the factual prompt with this cell marked obstacle, and $s_i^F$ the counterfactual prompt with the same cell marked free. Position, goal, previously observed cells, and other task
conditions are held fixed; the status of the edited cell is changed consistently in both the maintained map and local feedback. We retain pairs for which the edit changes the optimistic remaining distance of at least one other feasible action. The controlled prompt change does
not itself assume that only value information changes internally.

Let $\widehat{\boldsymbol{r}}(s)\in\mathbb{R}^{4}$ be the frozen readout's predicted remaining distances in the order
$(\mathrm{Up},\mathrm{Right},\mathrm{Down},\mathrm{Left})$, and define the centering matrix
$C=I_4-\frac14\boldsymbol{1}\boldsymbol{1}^{\top}$.
The implemented training differences are
\begin{equation}
\begin{aligned}
 \Delta\boldsymbol{v}_i
 &=C\bigl(\widehat{\boldsymbol{r}}(s_i^O)
          -\widehat{\boldsymbol{r}}(s_i^F)\bigr)
    \in\mathbb{R}^{4},\\
 \Delta\boldsymbol{h}_{\ell,i}
 &=\boldsymbol{h}_\ell(s_i^O)-\boldsymbol{h}_\ell(s_i^F)
    \in\mathbb{R}^{D_\ell}.
\end{aligned}
\label{eq:app_grid_fo}
\end{equation}
Thus, shortest-path targets identify informative F/O pairs, whereas NVT is fitted to \emph{decoded}, mean-centered distance differences and their matched prompt-last activation differences. Grid uses equal
weights $w_i=1$, with $B_\ell\in\mathbb{R}^{D_\ell\times4}$.

In \emph{WebShop}, a training comparison contains two candidate actions $a_i,a_j$ at the same branch state $s$. Define the scalar empirical continuation-reward difference and the corresponding candidate-span activation difference as
\begin{equation}
\begin{aligned}
 \Delta v_i
 &=\widehat Q^\pi(s,a_i)-\widehat Q^\pi(s,a_j)
    \in\mathbb{R},\\
 \Delta\boldsymbol{h}_{\ell,i}
 &=\boldsymbol{h}_\ell(s,a_i)-\boldsymbol{h}_\ell(s,a_j)
    \in\mathbb{R}^{D_\ell}.
\end{aligned}
\label{eq:app_webshop_pair}
\end{equation}
Here $\boldsymbol{h}_\ell(s,a)$ is the mean activation over the candidate's token span; $\widehat Q^\pi$ is estimated from sampled continuations as in Appendix~\ref{app:value-targets}. If a branch
state contributes $n_s$ pairs, each receives weight proportional to $1/n_s$. Consequently, $B_\ell\in\mathbb{R}^{D_\ell\times1}$,
which we identify with a vector in $\mathbb{R}^{D_\ell}$.

In either task, NVT maps a desired value-coordinate change to an edit:
\begin{equation}
 \boldsymbol{d}_\ell^{\mathrm{NVT}}
  =B_\ell\delta\boldsymbol{v}\in\mathbb{R}^{D_\ell}.
 \label{eq:app_nvt_edit}
\end{equation}
In Grid $\delta\boldsymbol{v}\in\mathbb{R}^4$; in WebShop
$\delta\boldsymbol{v}=(\delta v)\in\mathbb{R}^1$ contains a scalar command $\delta v$. The command comes from a fitted value readout, not held-out ground-truth value labels.

\paragraph{Grid value command and probe controls.}
Write $a^-$ for the original action and $a^+$ for the action preferred by the decoded remaining-distance ranking. For a unit positive dose,
Grid uses the centered distance command
\begin{equation}
 \delta\boldsymbol{v}
  =\tfrac12(\boldsymbol{e}_{a^-}-\boldsymbol{e}_{a^+})
  \in\mathbb{R}^{4},
 \label{eq:app_grid_command}
\end{equation}
where $\boldsymbol{e}_a$ is the one-hot action vector. This raises the decoded distance of $a^-$ and lowers that of $a^+$; multiplying the edit
by $\alpha$ supplies the desired dose. Let the layer-specific linear distance readout be
$\widehat{\boldsymbol{r}}_\ell(\boldsymbol{h})
 =W_\ell\boldsymbol{h}+\boldsymbol{b}_\ell$,
with $W_\ell\in\mathbb{R}^{4\times D_\ell}$ in raw activation coordinates. The calibrated probe intervention is the minimum-norm least-squares solution for the \emph{four-dimensional} decoded change:
\begin{equation}
 \boldsymbol{d}_\ell^{\mathrm{probe\text{-}cal}}
 =W_\ell^\top(W_\ell W_\ell^\top)^\dagger
   \delta\boldsymbol{v},
 \label{eq:app_grid_probe_cal}
\end{equation}
where $\dagger$ denotes the Moore--Penrose pseudoinverse. In particular, this construction is not restricted to changing only one scalar pairwise margin.

\paragraph{WebShop value command and probe controls.}
The WebShop pairwise readout is a logistic classifier, whose score must be calibrated to continuation-reward units. If $\boldsymbol{x}$ is a candidate-span activation difference, its score has the form
\begin{equation}
 z(\boldsymbol{x})
 =\boldsymbol{\theta}^{\top}
 \bigl((\boldsymbol{x}-\boldsymbol{\mu})\oslash
        \boldsymbol{\sigma}\bigr),
\end{equation}
where $\boldsymbol{\mu}$ and $\boldsymbol{\sigma}$ are training-set feature normalization statistics and $\oslash$ denotes element-wise division. A scalar training-only, out-of-fold regression gain $\gamma$
converts the score into the scalar command
$\delta v=\widehat{\Delta Q}=\gamma z(\boldsymbol{x})$.
For an intervention layer $\ell$, let
\begin{equation}
 \boldsymbol{u}_\ell
 =\frac{\boldsymbol{\theta}_\ell\oslash
             \boldsymbol{\sigma}_\ell}
        {\|\boldsymbol{\theta}_\ell\oslash
             \boldsymbol{\sigma}_\ell\|_2}
 \in\mathbb{R}^{D_\ell}
\end{equation}
be the unit direction of a training-fitted probe at that layer. Here the readout supplying $\delta v$ need not be the same probe as the layer-specific direction $\boldsymbol{u}_\ell$.
The WebShop calibrated probe edit retains only the component of NVT along that direction:
\begin{equation}
\begin{aligned}
 \boldsymbol{d}_\ell^{\mathrm{probe\text{-}cal}}
 &=\boldsymbol{u}_\ell\boldsymbol{u}_\ell^\top
      \boldsymbol{d}_\ell^{\mathrm{NVT}}\\
 &=\delta v\,(\boldsymbol{u}_\ell^\top B_\ell)
      \boldsymbol{u}_\ell.
\end{aligned}
\label{eq:app_webshop_probe_cal}
\end{equation}
For the symmetric pair intervention, the decoder-block output over the first candidate span receives
$+\alpha\boldsymbol{d}_\ell/2$, while the second receives
$-\alpha\boldsymbol{d}_\ell/2$. Hence the difference between their mean representations changes by $\alpha\boldsymbol{d}_\ell$.

\paragraph{Norm-matched and shuffled controls.}
For \emph{Probe (same-norm)}, orient the unit probe edit toward the action favored by the value command, and denote this unit vector by
$\widetilde{\boldsymbol{u}}_\ell$. The control is
\begin{equation}
 \boldsymbol{d}_\ell^{\mathrm{probe\text{-}norm}}
 =\|\boldsymbol{d}_\ell^{\mathrm{NVT}}\|_2
   \widetilde{\boldsymbol{u}}_\ell.
\end{equation}
In Grid the unit direction is obtained by normalizing
Eq.~\ref{eq:app_grid_probe_cal}; in WebShop it is the signed layer-specific probe direction. Zero-norm edits remain zero.

The \emph{Shuffled NVT} control is task-specific. In Grid, a random permutation $\sigma$ of training-pair indices breaks the correspondence between the four-dimensional value differences and their matched hidden-state differences. We refit the same objective
using $(\Delta\boldsymbol{v}_{\sigma(i)},
\Delta\boldsymbol{h}_{\ell,i})$ to obtain
$\widetilde B_\ell$. The shuffled edit
$\widetilde B_\ell\delta\boldsymbol{v}$ is then rescaled to the norm of $B_\ell\delta\boldsymbol{v}$.
In WebShop, the scalar-to-hidden transport $B_\ell$ is first fitted normally; a fixed permutation matrix
$P_\ell\in\mathbb{R}^{D_\ell\times D_\ell}$ then shuffles its hidden coordinates:
\begin{equation}
 \boldsymbol{d}_\ell^{\mathrm{shuf}}
 =P_\ell B_\ell\delta v,
 \qquad
 \|\boldsymbol{d}_\ell^{\mathrm{shuf}}\|_2
 =\|\boldsymbol{d}_\ell^{\mathrm{NVT}}\|_2.
\end{equation}
Thus, Grid shuffles the \emph{training alignment} before fitting the transport, whereas WebShop permutes the \emph{fitted transport} afterward; the WebShop operation is a coordinate permutation, not an arbitrary rotation.

For \emph{Random}, draw a nonzero isotropic Gaussian vector
$\boldsymbol{r}_\ell\sim\mathcal N(0,I_{D_\ell})$ and use
\begin{equation}
 \boldsymbol{d}_\ell^{\mathrm{rand}}
 =\|\boldsymbol{d}_\ell^{\mathrm{NVT}}\|_2
   \frac{\boldsymbol{r}_\ell}{\|\boldsymbol{r}_\ell\|_2}.
\end{equation}
The random direction is fixed across intervention strengths for a given evaluation instance and layer. Both controls separate the learned value-associated edit from intervention magnitude alone.

\paragraph{Policy margin and steering slope.}
For held-out comparison $i$ at state $s_i$, orient the action pair $(a_i^+,a_i^-)$ by the value-readout command. In Grid the selected comparisons also satisfy the corresponding task-value correctness criterion; in WebShop the command's sign need not agree with the
empirical reward ordering. The measured policy
margin is
\begin{equation}
 m_{\ell,i}(\alpha)
 =\log\pi_{\ell,\alpha}(a_i^+\mid s_i)
  -\log\pi_{\ell,\alpha}(a_i^-\mid s_i).
\end{equation}
WebShop evaluates this margin using exact action log probabilities; its pair is oriented by the sign of the calibrated readout estimate.
Define $\Delta m_{\ell,i}(\alpha)=m_{\ell,i}(\alpha)-m_{\ell,i}(0)$.
We fit, with an intercept, the OLS regression over
$\alpha\in\{0,1,2,4,8,16\}$:
\begin{equation}
 \Delta m_{\ell,i}(\alpha)
 =\kappa_{\ell,i}\alpha+\eta_{\ell,i}
   +\epsilon_{\ell,i}(\alpha).
\end{equation}
A positive $\kappa_{\ell,i}$ indicates an increasing preference for the readout-favored action. Grid reports the mean over evaluation states with group-clustered uncertainty. WebShop first averages pair-level slopes within each state, then equally across states, so states with more candidate pairs do not dominate the result.

\subsection{Closed-Loop Steering Settings}
\label{app:steering_settings}

Table~\ref{tab:steering_settings} lists the sensor layers,
NVT actuator layers, and intervention strengths used in
the behavioral experiments.

\begin{table}[t]
\centering
\small
\begin{tabular}{llccc}
\toprule
Task & Model & Sensor & Actuator & $\alpha$ \\
\midrule
Grid & Qwen 7B     & 19 & 21 & 12.5 \\
Grid & GPT-OSS 20B & 17 & 20 & 4 \\
Grid & Qwen 32B    & 51 & 50 & 12.5 \\
\midrule
WebShop & Qwen 7B     & 17 & 21 & 8 \\
WebShop & GPT-OSS 20B & 17 & 15 & 12.5 \\
WebShop & Qwen 32B    & 36 & 51 & 8 \\
\bottomrule
\end{tabular}
\caption{Configurations used for closed-loop value steering. Sensor and actuator refer to the value-readout and NVT intervention layers, respectively. Layer indices are zero-based.}
\label{tab:steering_settings}
\end{table}

\section{Prompt and Interaction Examples}
\label{app:prompts}

The two environments expose different forms of state to the agent. In Grid, we maintain an explicit sparse memory of all cells revealed so far and render this memory into the prompt at every decision step. The model therefore does not need to reconstruct the explored map from the raw interaction history. In WebShop, the agent instead follows the natural simulator trajectory: each decision is conditioned on the shopping instruction, the action prefix, the current page observation, and the actions currently available in the environment.

\paragraph{Reasoning configuration and activation extraction.}
Our prompts do not explicitly elicit reasoning traces, although the models differ in their native reasoning capabilities. Qwen2.5-7B-Instruct and Qwen2.5-32B-Instruct do not have a dedicated thinking mode, while GPT-OSS-20B is run with its native reasoning effort set to \texttt{low}. Importantly, all activations used for value probing and NVT are extracted from the input decision prompt, before any reasoning or action tokens are generated: at the final prompt token in Grid and over candidate-identifier spans in WebShop.

\subsection{Grid}

At each Grid step, the prompt contains the current position and goal, the maintained spatial memory, and fresh local feedback for the four neighboring
directions. The memory distinguishes confirmed free cells (\texttt{F}) and confirmed obstacles (\texttt{O}); all remaining in-bounds cells are treated as unknown. Fresh local feedback is incorporated into the maintained memory before the next decision.

The Grid system prompt is:
\begin{quote}
\small\ttfamily
You navigate a partially observed Cartesian grid.\\[2pt]
Coordinates are [x,y], with [0,0] at bottom-left.\\
UP increases y, DOWN decreases y, RIGHT increases x, and LEFT decreases x.\\[2pt]
You may move exactly one orthogonally adjacent grid cell per executed step.\\
Diagonal moves, jumping over cells, and staying in place are not allowed.\\[2pt]
Your goal is to reach the target in as few executed steps as possible.\\[2pt]
You are given a maintained spatial memory:\\
F = confirmed FREE: observed and traversable.\\
O = confirmed OBSTACLE: observed and not traversable.\\
All other in-bounds cells are UNKNOWN.\\
UNKNOWN does not mean free and does not mean obstacle.\\[2pt]
The spatial memory is cumulative and already incorporates all observations so far.\\
Do not reconstruct the map from dialogue history, and do not invent the state of\\
unknown cells.\\[2pt]
Choose the coordinate of the next cell you want to reach.\\
The next cell must be exactly one orthogonally adjacent cell away and must not be a\\
known obstacle or outside the grid.\\[2pt]
Return exactly one line and no explanation:\\
STEP1: [x,y]
\end{quote}

The following excerpt shows two consecutive decisions from a natural trajectory. At the first step, the cell directly above the agent has not yet been observed to be blocked:
\begin{quote}
\small\ttfamily
Grid size: 5x5\\
Current position: [1,0]\\
Goal: [1,4]\\[2pt]
Maintained spatial memory:\\
F: [0,0] [1,0] [2,0] [1,1] [1,4]\\
O: (none)\\[2pt]
Current local feedback:\\
UP [1,1]=F\\
RIGHT [2,0]=F\\
DOWN [1,-1]=OUT\\
LEFT [0,0]=F\\[2pt]
Model action: UP
\end{quote}

After moving upward, the environment reveals that \([1,2]\) is an obstacle.
The next prompt contains this new observation together with the previously revealed cells:
\begin{quote}
\small\ttfamily
Current position: [1,1]\\
Goal: [1,4]\\[2pt]
Maintained spatial memory:\\
F: [0,0] [1,0] [2,0] [0,1] [1,1] [2,1] [1,4]\\
O: [1,2]\\[2pt]
Current local feedback:\\
UP [1,2]=O\\
RIGHT [2,1]=F\\
DOWN [1,0]=F\\
LEFT [0,1]=F\\[2pt]
Model action: RIGHT
\end{quote}

Thus, the Grid prompt provides the model with a compact, cumulatively maintained representation of what has already been observed. This is also the partial map
used to define the optimistic prospective values in
Section~\ref{sec:value_analysis}: known obstacles are blocked, while unrevealed cells remain traversable.

\subsection{WebShop Full Natural Trajectory}
\label{app:webshop-full-trace}

For WebShop we show the complete natural trajectory of task 3, including the exact system prompt, every user prompt, the raw model output, the executed action, and the terminal state.

\paragraph{System prompt.}
\begin{quote}
\small\ttfamily
You operate a WebShop shopping environment.\\[2pt]
Your goal is to satisfy the user's shopping instruction and complete the purchase.\\[2pt]
Return exactly ONE action and nothing else.\\[2pt]
Allowed action formats:\\
search[query]\\
click[value]\\[2pt]
Rules:\\
- Use search[...] only when has\_search\_bar is true.\\
- For click[...], the value must exactly match one of the current clickables.\\
- On a search page, choose useful concise search terms from the instruction.\\
- On a results page, click the product ID that best matches the instruction.\\
- On a product page, select required options such as size or color before buying.\\
- When the product satisfies the request and required options are selected, click[buy now].\\
- Never explain your action.
\end{quote}

\paragraph{Step 0.}
\begin{quote}
\small\ttfamily
Instruction:\\
Instruction: Find me dual band streaming media players with quad core, and price lower than 340.00 dollars\\[2pt]
Trajectory so far:\\
(none)\\[2pt]
Current page:\\
WebShop [SEP] Instruction: [SEP] Find me dual band streaming media players with quad core, and price lower than 340.00 dollars [SEP] Search\\[2pt]
Allowed actions:\\
has\_search\_bar = True\\[2pt]
Allowed click targets:\\
- search\\[2pt]
For click[value], copy the target text exactly as shown above.\\
Do not add quoting or escape characters.\\[2pt]
Return exactly one action.
\end{quote}
\begin{quote}
\small\ttfamily
Raw model output:\\
search[dual band streaming media players quad core price < 340.00]\\[2pt]
Executed action:\\
search[dual band streaming media players quad core price < 340.00]\\[2pt]
reward\_after\_action: 0.0\\
done\_after\_action: False\\
state\_fingerprint\_before\_match: True\\
state\_fingerprint\_after\_match: True
\end{quote}

\paragraph{Step 1.}
\begin{quote}
\small\ttfamily
Instruction:\\
Instruction: Find me dual band streaming media players with quad core, and price lower than 340.00 dollars\\[2pt]
Trajectory so far:\\
- search[dual band streaming media players quad core price < 340.00]\\[2pt]
Current page:\\
Instruction: [SEP] Find me dual band streaming media players with quad core, and price lower than 340.00 dollars [SEP] Back to Search [SEP] Page 1 (Total results: 50) [SEP] Next > [SEP] B09LSKQF8C [SEP] Superbox S3 Pro Dual Band Wi-Fi 2.4Ghz 5Ghz Supports 6K Video [SEP] \$329.0 [SEP] B09NMZS7HK [SEP] EWEAT R10 II 4K UHD Media Player HDD Home Cinema VS10 Image Quality Engine HDR10+ 2GB+32GB Android Media Box with DAC Audio [SEP] \$599.0 [SEP] B00R249C5G [SEP] Remote for RCA Wi-fi Streaming Media Player DSB772E with 1080p HDMI output [SEP] \$49.95 [SEP] B07SRS3MXP [SEP] XTRONS 7 Inch Touch Screen Car Stereo for Jeep Wrangler Dodge Chrysler, Android 10 GPS Navigation for Car, Octa Core 4GB+64GB Car Radio Player, Built-in DSP Car Auto Play Android Auto Support 4G LTE [SEP] \$329.89 [SEP] B00OFQR8D2 [SEP] Visual Land Prestige Elite 10" Quad Core Tablet with KitKat 4.4, Google Play and Keyboard Bundle (Purple) [SEP] \$100.0 [SEP] B0871VQT3Q [SEP] Computer \& Tablet Y12 4G Phone Call Tablet PC, 10.1 inch, 2GB+32GB, Android 7.0 MTK6753 Octa-core up to 1.6GHz, WiFi, Bluetooth, OTG, GPS(Gold) (Color : Rose Gold) [SEP] \$100.0 [SEP] B085VFMXMZ [SEP] PRO OTG Adapter Works for LG V30 for OTG and USB Type-C Braided Cable. Use with Devices Like Keyboard, Mouse, Zip, Gamepad, hdmi, More (Gray) [SEP] \$7.99 [SEP] B01LZWDNX6 [SEP] VR Headset Homido V2 for iPhone and Android [SEP] \$54.99 [SEP] B08QSNM69H [SEP] 4K HDMI Cables, Short Cord, 1ft [3 Pack] Braided Cord \& Gold Connectors, 1ft High-Speed HDMI 2.0 with Ethernet, for Playstation PS5, PS4, PS3, Xbox, Roku, Apple TV, HDTV, Blu-ray, Laptop, PC, Monitor [SEP] \$15.49 [SEP] B08M5BYQN9 [SEP] MekedeTech 10.25'' Android 10 8 Core 4+64G Car Radio Multimedia Player GPS Navigation for Mercedes Benz S W221 W216 CL 2010-2013 S-Class Head Unit (NTG 3.5) [SEP] \$599.0\\[2pt]
Allowed actions:\\
has\_search\_bar = False\\[2pt]
Allowed click targets:\\
- back to search\\
- next >\\
- b09lskqf8c\\
- b09nmzs7hk\\
- b00r249c5g\\
- b07srs3mxp\\
- b00ofqr8d2\\
- b0871vqt3q\\
- b085vfmxmz\\
- b01lzwdnx6\\
- b08qsnm69h\\
- b08m5byqn9\\[2pt]
For click[value], copy the target text exactly as shown above.\\
Do not add quoting or escape characters.\\[2pt]
Return exactly one action.
\end{quote}
\begin{quote}
\small\ttfamily
Raw model output:\\
click[B09LSKQF8C]\\[2pt]
Executed action:\\
click[B09LSKQF8C]\\[2pt]
reward\_after\_action: 0.0\\
done\_after\_action: False\\
state\_fingerprint\_before\_match: True\\
state\_fingerprint\_after\_match: True
\end{quote}

\paragraph{Step 2.}
\begin{quote}
\small\ttfamily
Instruction:\\
Instruction: Find me dual band streaming media players with quad core, and price lower than 340.00 dollars\\[2pt]
Trajectory so far:\\
- search[dual band streaming media players quad core price < 340.00]\\
- click[B09LSKQF8C]\\[2pt]
Current page:\\
Instruction: [SEP] Find me dual band streaming media players with quad core, and price lower than 340.00 dollars [SEP] Back to Search [SEP] < Prev [SEP] Superbox S3 Pro Dual Band Wi-Fi 2.4Ghz 5Ghz Supports 6K Video [SEP] Price: \$329.0 [SEP] Rating: N.A. [SEP] Description [SEP] Features [SEP] Reviews [SEP] Buy Now\\[2pt]
Allowed actions:\\
has\_search\_bar = False\\[2pt]
Allowed click targets:\\
- back to search\\
- < prev\\
- description\\
- features\\
- reviews\\
- buy now\\[2pt]
For click[value], copy the target text exactly as shown above.\\
Do not add quoting or escape characters.\\[2pt]
Return exactly one action.
\end{quote}
\begin{quote}
\small\ttfamily
Raw model output:\\
click[buy now]\\[2pt]
Executed action:\\
click[buy now]\\[2pt]
reward\_after\_action: 1.0\\
done\_after\_action: True\\
state\_fingerprint\_before\_match: True\\
state\_fingerprint\_after\_match: True
\end{quote}

\paragraph{Terminal state.}
\begin{quote}
\small\ttfamily
Thank you for shopping with us! [SEP] Your code: [SEP] None [SEP] (Paste it in your MTurk interface.) [SEP] Purchased [SEP] asin [SEP] B09LSKQF8C [SEP] options [SEP] \{\} [SEP] attrs [SEP] None [SEP] category [SEP] None [SEP] query [SEP] None [SEP] product category [SEP] None [SEP] Target [SEP] asin [SEP] options [SEP] attrs [SEP] price upper [SEP] instuction text [SEP] category [SEP] product category [SEP] query [SEP] Goal [SEP] None [SEP] Reward [SEP] Your score (min 0.0, max 1.0) [SEP] 1.0 [SEP] Reward Details [SEP] None\\[2pt]
terminal\_prefix\_actions: ["search[dual band streaming media players quad core price < 340.00]", "click[B09LSKQF8C]", "click[buy now]"]\\
terminal\_reward: 1.0\\
terminal\_done: True
\end{quote}

\section{Attributions and Licenses}
\label{app:attributions}

\paragraph{Models.}
We use three publicly available language models:
Qwen2.5-7B-Instruct
(\url{https://huggingface.co/Qwen/Qwen2.5-7B-Instruct}),
Qwen2.5-32B-Instruct
(\url{https://huggingface.co/Qwen/Qwen2.5-32B-Instruct}),
and GPT-OSS-20B
(\url{https://huggingface.co/openai/gpt-oss-20b}).
All three model checkpoints are released under the
Apache 2.0 license.

\paragraph{Environment.}
We use the WebShop environment
\citep{NEURIPS2022_82ad13ec}, available at
\url{https://github.com/princeton-nlp/WebShop}.
The official WebShop implementation is released
under the MIT license.

\paragraph{Compute resources.}
All experiments were conducted using four
NVIDIA H100 GPUs, requiring approximately 200 GPU-hours in total. 



\end{document}